\documentclass[english]{llncs}

\usepackage{latexsym}
\usepackage{float}
\usepackage{amssymb}
\usepackage{amsmath}
\usepackage{url}

\newcommand{\cmdline}[1]{\texttt{\textbf{#1}}}

\newcommand{\plc}{\mathit{PL}^{cc}}
\newcommand{\psp}{\mathit{PS}^{+}}

\newcommand{\wsat}{\mathit{WSAT}}
\newcommand{\smd}{\mathit{smodels}}
\newcommand{\wsatoip}{\mathit{WSAT(OIP)}}

\newcommand{\was}{\it{WSAT(CC)}}

\newcommand{\vwcc}{\mathit{WSAT(CC)\mbox{-}VB}}
\newcommand{\dwcc}{\mathit{WSAT(CC)\mbox{-}DF}}
\newcommand{\pwcc}{\mathit{WSAT(CC)\mbox{-}PF}}

\floatstyle{boxed}
\newfloat{algorithm}{tbp}{loa}
\floatname{algorithm}{Algorithm}

\numberwithin{equation}{section} 
\numberwithin{figure}{section} 

\usepackage{babel}
\makeatother
\begin{document}

\title{$\was$ --- a fast local-search ASP solver}
\titlerunning{$\was$ a local-search ASP solver}

\author{Lengning Liu \and Miros\l aw Truszczy\'nski}
\authorrunning{Lengning Liu and Miros\l aw Truszczy\'nski}
\institute{Department of Computer Science, University
  of Kentucky, Lexington, KY 40506-0046, USA}

\maketitle

\begin{abstract}
We describe $\was$, a local-search solver for computing models of
theories in the language of propositional logic extended by
cardinality atoms. $\was$ is a processing back-end for the logic
$\psp$, a recently proposed formalism for answer-set programming.
\end{abstract}

\section{Introduction}
\vspace*{-0.1in}

$\was$ is a local-search solver for computing models of theories
in the logic $\plc$, the propositional logic extended by cardinality
atoms \cite{et01a,lt03}. It can serve as a processing back-end for
the logic $\psp$ \cite{et01a}, an {\em answer-set programming} (ASP)
formalism based on the language of predicate calculus and, hence, 
different from typical ASP systems that have origins in logic programming.

A {\em clause} in the logic $\plc$ is a formula $\alpha_1 \wedge
\ldots \wedge \alpha_r \rightarrow \alpha_{r+1} \vee \ldots \vee
\alpha_s$, where each $\alpha_i$, $1\leq i\leq s$, is a propositional
atom, or a {\em cardinality atom} ({\em c-atom}, for short) --- an
expression $k\{a_1, \ldots, a_n \}m$, where $a_i$ are propositional
atoms and $k,m$ and $n$ are integers such that $0\leq k\leq m\leq n$.
A set of atoms $M$ is a {\em model} of a c-atom $k\{a_1,\ldots,a_n\}m$
if $k\leq |M\cap \{a_1,\ldots,a_n\}|\leq m$. With this definition,
the semantics of clauses and theories in the logic $\plc$ is a
straightforward extension of the semantics of propositional logic.
$\plc$ theories arise by grounding theories in the ASP logic $\psp$
by means of the grounder program {\em asppsgrnd} \cite{et01a}.

We discuss here an implementation of $\was$. We restrict the discussion
to most essential concepts and options only. For more details
and bibliography on related work on propositional logic extended by
c-atoms and pseudo-boolean constraints, we refer
to \cite{lt03}, which introduced $\was$, and to \cite{aspps-users-man}.

\section{$\was$ --- a brief description and a list of options}
\vspace*{-0.1in}

As other WSAT-like local-search solvers \cite{skc94,wal97}, $\was$
searches for models in a series of {\em tries}, starting with
a random assignment of truth values to atoms. Each try consists of steps,
called {\em flips}, which produce "new" truth assignments by {\em flipping}
the truth values of some of the atoms.  If a flip produces a satisfying
assignment, this try is terminated and another one starts. $\was$ supports
several strategies to select atoms for flipping. All of them require a
parameter called the {\em noise level}. It determines the probability
of applying a random walk step in order to escape from 
a local minimum. The maximum numbers of tries and flips, and the noise 
level are set from the command line by means of the options
{\tt -t}, {\tt -c} and {\tt -N}, respectively.

$\was$ is different from other similar algorithms in the way
in which it computes the {\em break-count} of an atom (used to decide
which atom to flip) and in the way it executes a flip. The choice of 
the break-count computation method or of the way a flip is defined 
determines a particular local-search strategy in $\was$. At present,
$\was$ supports three basic methods.\\
{\bf Virtual break-count.} We define virtual break-counts with
respect to a propositional theory, in which c-atoms are replaced by
their equivalent propositional representations. However, in the actual
computation we use the original theory (with c-atoms) rather than
its propositional-logic counterpart (with c-atoms removed), as the
latter is usually exponentially larger. To invoke virtual break-count 
method, we use the option {\tt -VB}. The virtual break-count method
is applicable with all $\plc$ theories and is a default method
of $\was$.
\\
{\bf Double flip.} It applies only to {\em simple} $\plc$ theories that 
are specified by the following two conditions: (a) all the c-atoms 
appear in unit clauses, and (b) all the sets of atoms in the c-atoms
are pairwise disjoint. A flip is designed so that all unit clauses 
built of c-atoms remain satisfied. Thus, on occasion, two atoms will 
change their truth values in one flip step. The break-count is defined
with respect to regular propositional clauses as in $\wsat$. To invoke 
this method, we use the option
{\tt -DF}.\\
{\bf Permutation flip.} It applies to theories, in which c-atoms are
used solely to specify permutations (for instance, when defining an 
assignment of queens in the $n$-queens problem). Flips realize
an inverse operation on permutations and, hence, transform 
a permutation into another permutation. As a consequence, all unit 
clauses built of c-atoms are always satisfied. To accomplish that, four 
atoms must have their truth values changed in one flip step. The
break-count is defined with respect to regular propositional clauses of
the theory in the same way as in $\wsat$. We invoke this method with 
the option {\tt -PF}.

\vspace*{-0.1in}
\section{$\was$ --- input, output and how to invoke it}
\vspace*{-0.1in}

$\was$ accepts input files containing $\plc$ theories described in a
format patterned after that of CNF DIMACS. The first line is of the form
{\tt p <na> <nc>}, where {\tt na} and {\tt nc} are the number of
propositional atoms and clauses in the theory, respectively. The
following lines list clauses. A clause $\alpha_1 \wedge \ldots \wedge
\alpha_r \rightarrow \alpha_{r+1} \vee \ldots \vee \alpha_s$, is
written as {\tt A1 ... Ar , A(r+1) ... As}, where each {\tt Ai} is
a positive integer (representing the corresponding atom $\alpha_i$),
or an expression of the form {\tt \{k m C1 ... Cn\}} (representing a
c-atom $k\{a_1,\ldots,a_n\}m$).

$\was$ outputs models that it finds as well as several statistics
to standard output device (or, depending on the options used, to a
file in a user-readable format). It also creates a file
\cmdline{wsatcc.stat} that stores records summarizing every call
to $\was$ and key statistics pertaining to the computation.

Typical call to $\was$ looks as follows: {\tt wsatcc -f file -t 200 -c
150000 -N 10 100}. It results in $\was$ looking for models to the $\plc$
theory specified in {\tt file}, by running 200 tries, each consisting of
150000 flips. The noise level is set at 10/100 (=0.1).

\vspace*{-0.1in}
\section{$\was$ package}
\vspace*{-0.1in}

$\was$ solver and several related utilities can be obtained from
\url{http://www.cs.uky.edu/ai/wsatcc/}. $\was$ works on most
Unix-like operating systems that provide \cmdline{gcc} compiler.
The utilities require Perl 5 or greater. For more details on
installation, we refer to \cite{aspps-users-man}.

\vspace*{-0.1in}
\section{Performance}
\vspace*{-0.1in}

Our experiments demonstrate that $\was$ is an effective tool to compute
models of {\em satisfiable} $\plc$ theories and can be used as a
processing back-end for the ASP logic $\psp$. In \cite{lt03}, we showed
that $\was$ is often much faster than a local-search SAT solver $\wsat$
and has, in general, a higher success rate (likelihood that it will
find a model if an input theory has one). In \cite{dmt03a}, we used
$\was$ to compute several new lower bounds for van der Waerden numbers.
Here, we will discuss our recent comparisons of $\was$ with $\wsatoip$
\cite{wal97}, a solver for propositional theories extended with
pseudo-boolean constraints (for which we developed utilities allowing it
to accept $\plc$ theories).

We tested these programs on $\plc$ theories
encoding instances of the vertex-cover 
and {\em open $n$-queens} problems\footnote{In the open $n$-queens
problem, given an initial ``attack-free'' assignment of $k$ ($k<n$)
queens on the $n\times n$ board, the goal is to assign the remaining
$n-k$ queens so that the resulting assignment is also ``attack-free''.}.
We generated these theories
by grounding appropriate $\psp$ theories
extended with randomly generated problem instances.

Table \ref{tab1} shows results obtained by running $\was$ (both {\em -VB}
and {\em -DF} versions are applicable in this case) and $\wsatoip$
to find vertex covers of sizes 1035, 1040 and 1045 in graphs with 2000
vertices and 4000 edges.
The first column shows the size of the desired vertex
cover and the number of graphs (out of 50 that we generated), for which
we were able to find a solution by means of at least one of the methods
used. The remaining columns summarize the performance of the three
algorithms used: $\vwcc$, $\dwcc$, and $\wsatoip$. The
entries show the {\em time}, in seconds, needed to complete computation for
all 50 instances and the {\em success rate} (the percentage of cases where the
method finds a solution to all the instances, for which at least one
method found a solution).

\vspace*{-0.1in}
\begin{table}[h]
{\scriptsize
\caption{Vertex cover: Large Graphs }
\label{tab1}
\vspace*{-0.1in}
\begin{center}
\begin{tabular}{|l|r|r|r|}
\hline
Family & $\vwcc$ & $\dwcc$ & $\wsatoip$ \\
\hline
\hline
{\em 1035 (9 / 50)} & 1453/77\% & 3426/100\% & 9748/11\% \\
\hline
{\em 1040 (24 / 50)} & 1166/95\% & 2464/100\% & 7551/100\%  \\
\hline
{\em 1045 (36 / 50)} & 991/86\% & 1610/100\% & 6365/100\% \\
\hline
\end{tabular}
\end{center}
}
\end{table}

\vspace*{-0.3in}
The results show that $\vwcc$ is faster than $\dwcc$,
which in turn is faster that $\wsatoip$. However, $\vwcc$ has
generally the lowest success rate while $\dwcc$, the highest.

We note that we attempted to compare $\was$ with $\smd$
\cite{ns00}, a leading ASP system. We found that for the large
instances that we experimented with $\smd$ failed to terminate within
the time limit that we allocated per instance. That is not surprising,
as the search space is prohibitively large for a complete method and
$\smd$ is a complete solver.

The open $n$-queens problem allowed us to experiment with the method
{\em -PF} (permutation flip). It proved extremely effective. We tested
it for the case of 50 queens with 10 of them preassigned. We generated
100 random preassignments of 10 queens to a $50\times 50$ board and
found that 55 of them are satisfiable. We tested the four algorithms
only on those satisfiable instances. The results are shown in
Table \ref{tab2}.

\vspace*{-0.1in}
\begin{table}[h]
{\scriptsize
\caption{Open $n$-Queens: $N=50$, $10$ preassigned}
\label{tab2}
\vspace*{-0.1in}
\begin{center}
\begin{tabular}{|l|r|r|r|r|}
\hline
Family & $\vwcc$ & $\pwcc$ & $\wsatoip$ & $\smd$ \\
\hline
\hline
{\em 50+10(55 / 55)} & 20/1539/100\% & 9/768/100\% & 76/1459/100\% & 908/10\% \\
\hline
\end{tabular}
\end{center}
}
\end{table}

\vspace*{-0.3in}
Here, we include another measurement for local search solvers. The
second number shows the average number of flips each method uses in finding one
solution. $\vwcc$ is faster than $\wsatoip$ even though
they have the similar number of flips. $\pwcc$ is even more powerful because it
uses the fewest number of flip and is the fastest. Smodels can
only find solutions for 6 instances within the 1000-second limit and turns out
to be the slowest.

We tested the version {\em -PF} with one of the encodings of the
Hamiltonian-cycle problem and discovered it is much less effective
there. Conditions under which the version {\em -PF} is effective remain
to be studied.

\vspace*{-0.1in}
\section*{Acknowledgments}
\vspace*{-0.1in}

This research was supported by the National Science
Foundation under Grants No. 0097278 and 0325063.


\begin{thebibliography}{1}

\bibitem{dmt03a}
M.R. Dransfield, V.M. Marek, and M.~Truszczy{\'n}ski.
\newblock Satisfiability and computing van der Waerden numbers.
\newblock In {\em Proceedings of SAT-2003}. LNAI, Springer Verlag, 2003.

\bibitem{aspps-users-man}
D.~East, L.~Liu, S.~Logsdon, V.~Marek, and M.~Truszczy\'nski.
\newblock {ASPPS} user's manual, 2003.
\newblock \url{http://www.cs.uky.edu/aspps/users_manual.ps}.

\bibitem{et01a}
D.~East and M.~Truszczy{\'n}ski.
\newblock Propositional satisfiability in answer-set programming.
\newblock In {\em Proceedings of KI-2001}, LNAI 2174. Springer Verlag, 2001.
\newblock Full version submitted for publication (available at
  \url{http://xxx.lanl.gov/abs/cs.LO/0211033}).

\bibitem{lt03}
L.~Liu and M.~Truszczy{\'n}ski.
\newblock Local-search techniques in propositional logic extended with
  cardinality atoms.
\newblock In {\em Proceedings of CP-2003}. LNCS, Springer Verlag, 2003.

\bibitem{ns00}
I.~Niemel{\"a} and P.~Simons.
\newblock Extending the smodels system with cardinality and weight constraints.
\newblock In J.~Minker, editor, {\em Logic-Based Artificial Intelligence},
  pages 491--521. Kluwer Academic Publishers, 2000.

\bibitem{skc94}
B.~Selman, H.A. Kautz, and B.~Cohen.
\newblock Noise strategies for improving local search.
\newblock In {\em Proceedings of AAAI-94}. AAAI Press, 1994.

\bibitem{wal97}
J.P. Walser.
\newblock Solving linear pseudo-boolean constraints with local search.
\newblock In {\em Proceedings of AAAI-97}. AAAI Press, 1997.

\end{thebibliography}

\end{document}